\documentclass{article}

\usepackage[main, final]{neurips_2026}

\usepackage[utf8]{inputenc} 
\usepackage[T1]{fontenc}    
\usepackage{hyperref}       
\usepackage{url}            
\usepackage{booktabs}       
\usepackage{amsmath}        
\usepackage{amsfonts}       
\usepackage{nicefrac}       
\usepackage{microtype}      
\usepackage{xcolor}         
\usepackage{multirow}       
\usepackage{graphicx}       
\graphicspath{{assets/}{figures/}{./}}
\usepackage{tikz}
\usetikzlibrary{positioning, arrows.meta, calc}
\usepackage{colortbl}       
\usepackage[capitalise,noabbrev]{cleveref} 
\usepackage{subcaption}     

\newcommand{\modelname}{OneCanvas}
\title{OneCanvas: 3D Scene Understanding \\ via Panoramic Reprojection}

\author{%
  Bart{\l}omiej Baranowski\textsuperscript{1} \quad
  Dave Zhenyu Chen\textsuperscript{2} \quad
  Matthias Nie{\ss}ner\textsuperscript{1} \\[0.4em]
  \textsuperscript{1}Technical University of Munich \quad
  \textsuperscript{2}Huawei
}

\begin{document}

\maketitle

\begin{center}
  \vspace{-0.5em}
  Project page: \url{https://baranowskibrt.github.io/onecanvas/}\\
  Code and model: \url{https://github.com/baranowskibrt/onecanvas}
\end{center}

\begin{figure}[h]
  \centering
  \includegraphics[width=\linewidth,trim=17 2 17 2,clip]{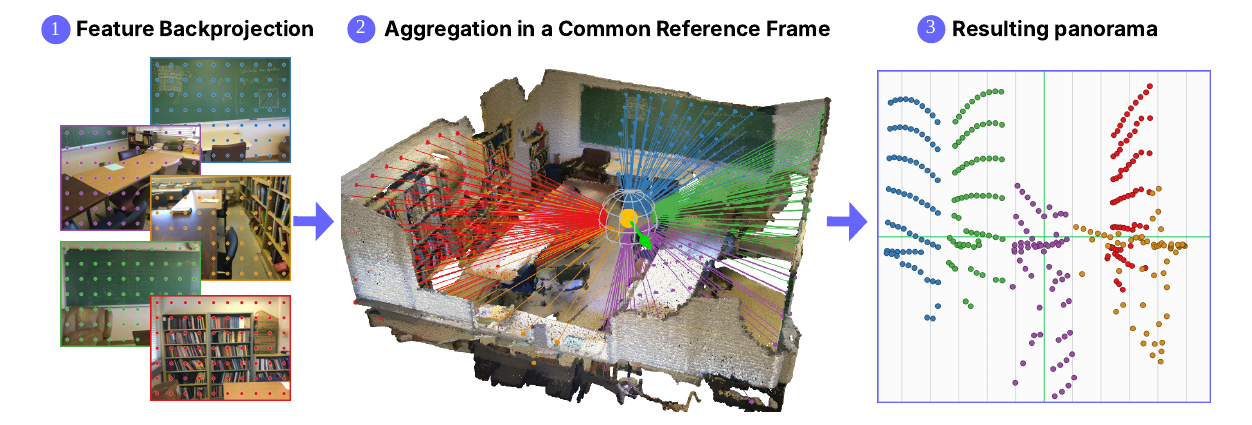}
  \caption{\textbf{\modelname{}} \textbf{(1)} backprojects per-frame patch features to 3D, \textbf{(2)} aggregates them in a common reference frame centered on a chosen viewpoint, and \textbf{(3)} places each patch at its continuous angular position on a panoramic canvas consumed by the VLM as one image.}
  \label{fig:teaser_image}
\end{figure}

\begin{abstract}

Existing approaches to 3D scene understanding in Vision-Language Models (VLMs) either rely on complex, model-specific geometry encoders or large training budgets in pursuit of spatial reasoning. Instead, \modelname{} aggregates patch features from all views onto a single equirectangular panoramic canvas. Namely, each patch is unprojected to a 3D world coordinate using its depth and camera pose, then placed on the canvas at the continuous longitude and latitude of that point as seen from the canvas origin, with no rasterization or aggregation across overlapping views. A 3D position embedding of the patch's metric coordinates is added to its feature, restoring the depth lost when collapsing the world position to an angular canvas coordinate. Patches from all frames thus share one spatial coordinate system with no fusion or major architectural modifications of the backbone. The pretrained VLM consumes this representation as if it were an ordinary image. Because the canvas can be centered on any pose of interest, the same representation directly supports situated reasoning from a specific viewpoint, a common requirement in robotics and embodied AI. Thanks to this representation, we can also introduce a spatial pretraining curriculum: by procedurally placing patch features of objects, drawn from real images, at chosen 3D world positions on an otherwise empty canvas, we generate on-the-fly supervision spanning a broad range of spatial reasoning tasks, with answer distributions controlled to reduce spatial reasoning shortcuts. \modelname{} achieves state-of-the-art results on SQA3D and VSI-Bench and generalizes to out-of-distribution data on SPBench, using an order of magnitude less training compute than its competitors closest in benchmark performance.

\end{abstract}

\begin{figure}[!t]
    \centering
    \includegraphics[width=\textwidth,trim=6 4 4 4,clip]{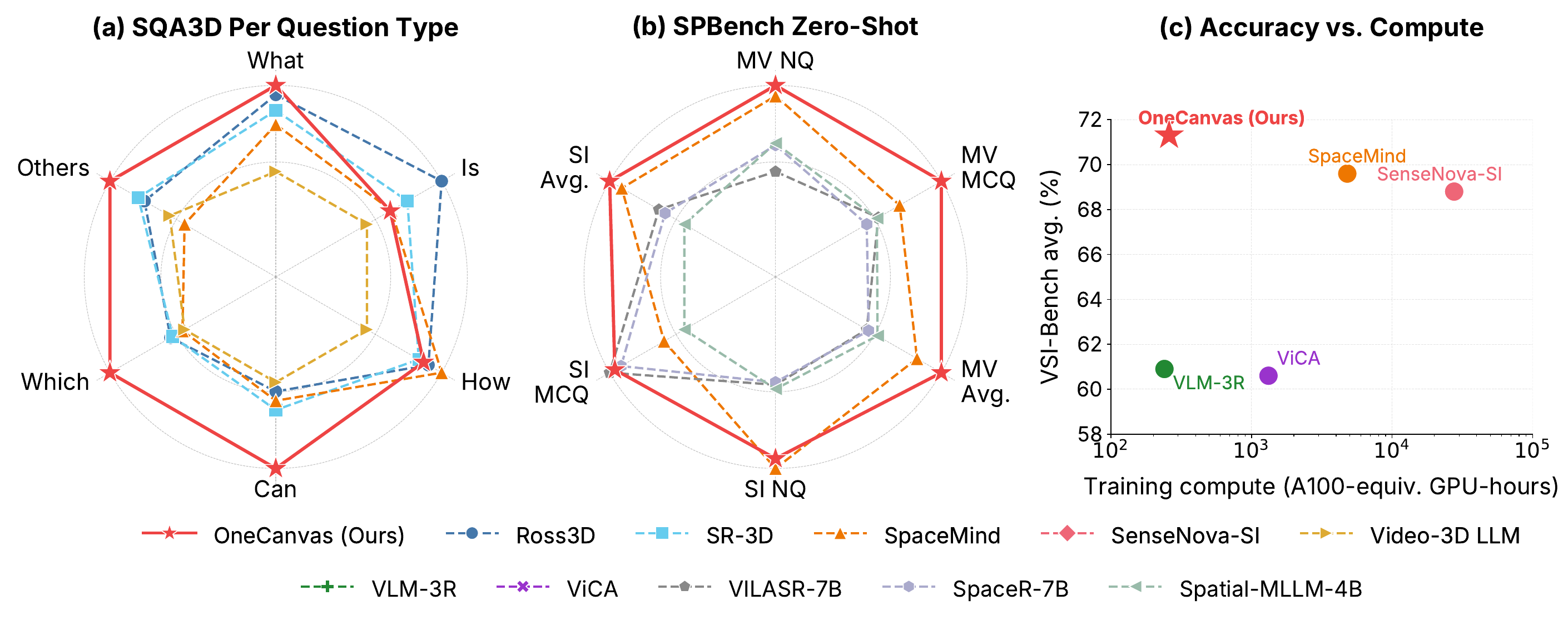}
    \caption{\textbf{\modelname{} benchmark comparison.}
    \textbf{(a)} Per-question-type accuracy on SQA3D and \textbf{(b)} zero-shot accuracy on SPBench
    (multi-view and single-image splits), shown as radar charts against competitors closest in benchmark performance.
    \textbf{(c)} Training compute (A100-equivalent GPU-hours, log scale) versus VSI-Bench accuracy.}
    \label{fig:benchmark_comparison}
\end{figure}

\section{Introduction}
\label{sec:intro}

Vision-language models have transformed how machines interpret images, video, and natural language jointly, but a great deal of practically valuable understanding is intrinsically three-dimensional. Robots that act in human environments, augmented-reality assistants that ground instructions in a user's physical surroundings, and autonomous agents that plan trajectories through unfamiliar spaces all need to answer questions about \emph{where} things are, how far apart they are, which direction one must turn to face them, and what is or is not visible from a given vantage point. Building VLMs that answer such questions reliably from ordinary video of a scene is the goal of 3D scene understanding.

Most existing 3D-aware VLMs are built on top of pretrained 2D vision-language backbones with no native machinery for spatial reasoning. Adapting them to 3D therefore demands either inducing geometry implicitly from raw video, which is data-inefficient and tends to confound geometry with scene-specific texture priors, or augmenting the backbone with a dedicated geometry encoder. Even with these adaptations in place, the supplied geometry is often not actually used. Recent audits on situated 3D question answering find that text-only baselines match or approach state-of-the-art geometry-aware models on widely cited benchmarks~\cite{ma2026real3dqa}, indicating that the models fall back on statistical regularities of the question and scene class rather than reasoning over the geometric input they are given.

Existing 3D VLMs have nonetheless made substantial progress on 3D reasoning. One family augments the VLM with a dedicated geometric module, including point-cloud tokenisers~\cite{ll3da,leo,chat3d,pq3d}, depth-conditioned video position encodings~\cite{video3dllm,llava3d}, and fused geometry encoders such as VLM-3R~\cite{vlm3r} and SpaceMind~\cite{spacemind} that read 3D structure from monocular video and inject it into the visual backbone. A second family scales spatial supervision: SenseNova-SI~\cite{sensenovasi} and Cambrian-S~\cite{cambrians} curate large collections of spatial QA pairs to teach a geometrically vanilla VLM to reason about distance and direction by sheer exposure. Both directions deliver concrete improvements on every major spatial benchmark, and together they show that 3D reasoning is reachable from a pretrained VLM if one is willing to invest in either architecture or data.

Both directions also pay a heavy bill. Architectural fusion adds modules that must be jointly trained at scale to align a new geometry encoder with the language backbone, while data scaling demands curating and sorting millions of high-quality spatial question-answer pairs. Neither directly closes the gap flagged by the audits above: methods in both families still see the scene only piecewise across frames and have no clear handle on which signals in the input actually encode the spatial relationships they are being asked to learn. Instead, we propose \modelname{} (\Cref{fig:teaser_image}): each patch from each view is lifted to a 3D world coordinate using its depth and camera pose, then placed on a single equirectangular canvas at the longitude and latitude of that point as seen from a chosen canvas origin. These spherical coordinates fill the spatial axes of the patch's MRoPE position, while the source-frame index occupies the temporal axis. All patches from all views thus share one spatial coordinate system, and the pretrained VLM consumes the result as if it were an ordinary single-image input, with no major architectural modifications. This representation also enables a spatial pretraining curriculum that would be difficult to assemble in conventional 2D or video pipelines. By procedurally placing patch features of objects, drawn from real images, at chosen 3D world positions on an otherwise empty canvas, we generate on-the-fly supervision for tasks ranging from metric distances to directions, counting, observability, and navigation. Answer distributions are controlled to suppress the scene-statistic shortcuts recent audits have flagged~\cite{ma2026real3dqa}. \modelname{} reaches state of the art on SQA3D~\cite{sqa3d} (65.3 EM@1, 2.3 points above the previous best), VSI-Bench~\cite{vsibench} (71.3 average), and SPBench~\cite{spbench} (72.1 zero-shot overall, 4.8 points above the next best method), while using an order of magnitude less training compute than its competitors closest in benchmark performance (\Cref{fig:benchmark_comparison}, with the full normalization methodology in \Cref{app:compute}).

Our contributions are:
\begin{itemize}
    \item \textbf{Panoramic feature reprojection.} A mechanism that aggregates patch features from all views onto a single panoramic canvas, matching the input format the VLM was pretrained on, allowing for flexible choice of canvas origin and orientation to suit different tasks.

    \item \textbf{Spatial pretraining curriculum enabled by the representation.} On-the-fly geometric supervision built by procedurally placing patch features of objects, drawn from real images, at chosen 3D positions on the canvas, with answer distributions controlled to reduce the dataset-statistic shortcuts.
\end{itemize}

\section{Related Work}
\label{sec:related}

\paragraph{3D-aware VLMs via architectural extensions.}
Most existing 3D VLMs equip a pretrained language backbone with auxiliary modules that supply geometric information~\cite{whenllms, ll3da, leo, chat3d, pq3d, uni3dl, video3dllm, vlm3r, spacemind, ross3d, sr3d, gpt4scene}.
Early point-cloud-based 3D VLMs~\cite{3dllm,ll3da,leo,chat3d,pq3d,uni3dl} pair a language backbone with a dedicated point cloud tokeniser and train end-to-end on 3D QA data, establishing the feasibility of LLM-based 3D scene reasoning.
RGB-D-based 3D VLMs~\cite{video3dllm,llava3d} instead treat the input as a video sequence and inject 3D position encodings computed from depth maps into the video representation, bridging the video and 3D domains without an explicit point-cloud encoder.
HiSpatial~\cite{liang2026hispatial} extends this line by feeding metric-scale point maps as auxiliary inputs alongside RGB-D and supervising the VLM through a hierarchical curriculum from geometric perception to abstract spatial reasoning.
With the emergence of geometry foundation models~\cite{dust3r,mast3r,vggt,depthanything3}, more recent 3D VLMs extract 3D tokens directly from monocular video and fuse them with visual features before the LLM~\cite{vlm3r,spacemind,ross3d,sr3d,gpt4scene,vgllm}.
VLM-3R~\cite{vlm3r} fuses implicit 3D tokens from a geometry encoder with visual features.
SpaceMind~\cite{spacemind} adds a Camera-Guided Modality Fusion module that combines a spatial encoder with a visual encoder.
Ross3D~\cite{ross3d} introduces a reconstructive visual instruction-tuning objective that supplies 3D-aware supervision during fine-tuning of a video VLM.
SR-3D~\cite{sr3d} is a region-promptable VLM that unifies single-view 2D and multi-view 3D inputs in a shared visual token space.
GPT4Scene~\cite{gpt4scene} fine-tunes Qwen2-VL on video sequences of 3D scenes.
SD-VLM~\cite{sdvlm} introduces a sinusoidal depth positional encoding that adds depth information along the camera's $z$-axis.
Loc3R-VLM~\cite{qu2026loc3rvlm} pairs camera priors from a pretrained geometry foundation model with auxiliary BEV-layout and situation-modeling objectives.
A parallel line uses reinforcement learning with spatial reward signals~\cite{spacer,vilasr,spatialthinker,thinkingspatialcode}, which is orthogonal to representation choice and could in principle be combined with our canvas.
All of these methods introduce new architectural components, auxiliary objectives, or dedicated 3D encoders on top of the base VLM.
In contrast, our approach introduces no new architectural components, no auxiliary loss terms, and no dedicated 3D encoder. 3D understanding emerges from how the input is prepared.

\paragraph{Unified scene representations.}
Panoramic image representations offer a 360$^\circ$ field of view that naturally preserves long-range spatial relationships, a property exploited for scene understanding~\cite{panorama_survey} and recently for 3D visual grounding.
PanoGrounder~\cite{panogrounder} renders multiple equirectangular panoramas from virtual cameras placed around a pre-reconstructed mesh or 3D Gaussian Splatting scene and processes them with a large VLM for grounding.
PanoEnv~\cite{panoenv} studies VLM spatial intelligence on equirectangular images from synthetic environments, showing that geometric distortion in ERP images remains a challenge for off-the-shelf VLMs.
The broader idea of unprojecting 2D features into 3D and reprojecting them onto a canonical 2D view is well-established in autonomous driving through BEV methods~\cite{lss,bev_survey}.
Lift3D~\cite{lift3d} lifts per-view features into a neural field and renders target views, and LiftProj~\cite{liftproj} uses a similar lift-then-project paradigm for panorama stitching.
Our work brings this principle to VLM-based 3D scene understanding: per-view features are lifted with metric depth and camera poses, projected once onto a panoramic canvas, and consumed directly by the VLM, with the canvas origin chosen freely to support both scene-centered and pose-centered situated reasoning.
Unlike PanoGrounder, LiftProj, and PanoEnv, we do not render an equirectangular image for a vision tower to re-encode: the tower sees the original perspective frames, and only the resulting patch positions are reparameterized. Unlike rendering or Lift3D's ray compositing, the canvas also retains occluded and co-located observations as separate tokens. Controlled comparisons are in \Cref{app:representation_controls}.

\section{Method}
\label{sec:method}

\begin{figure*}[t]
    \centering
    \includegraphics[width=\textwidth,trim=19 6 20 0,clip]{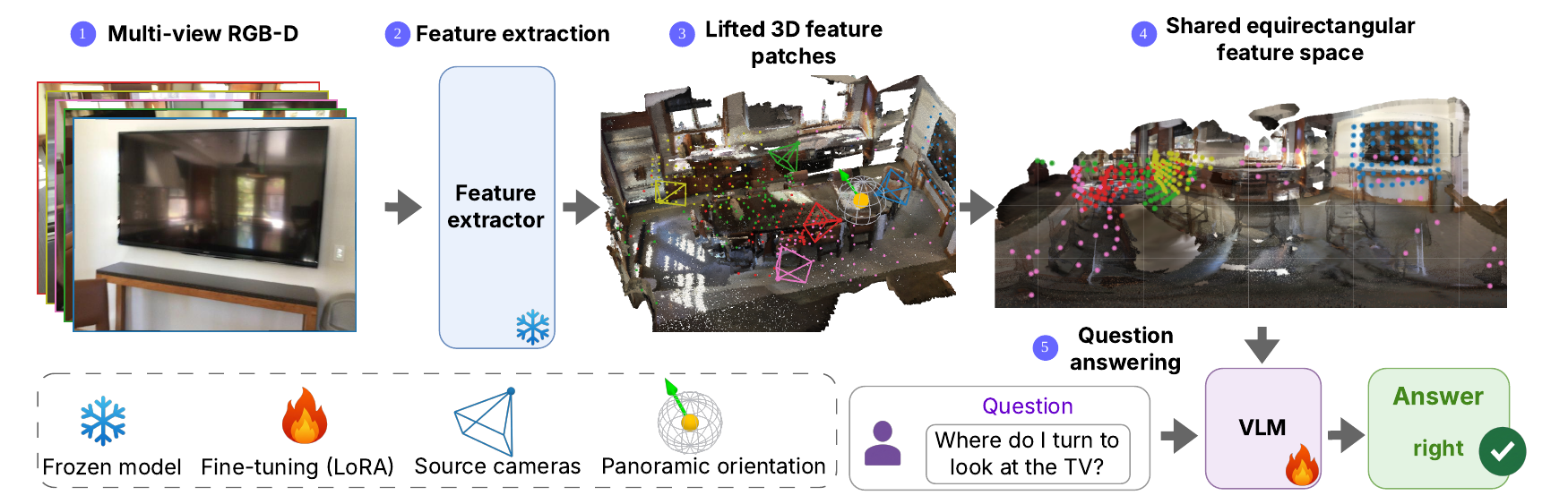}
    \caption{\textbf{Method overview.} Multi-view images are encoded by Qwen3-VL's frozen vision encoder, and lifted patches land at continuous positions on a shared equirectangular panorama. The VLM consumes this representation through its native attention layers.}
    \label{fig:method}
\end{figure*}

\begin{figure}[t]
    \centering
    \includegraphics[width=\linewidth,trim=17 2 17 2,clip]{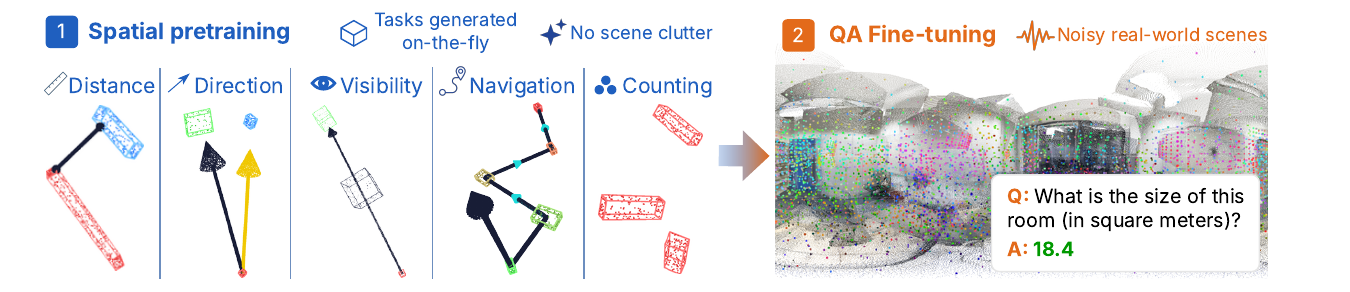}
    \caption{\textbf{Two-stage training.} Stage 1 trains LoRA~\cite{lora} adapters and the 3D position embedding on the spatial pretraining curriculum, where objects placed on an otherwise empty canvas and their inline references carry the geometric signal. Stage 2 merges the stage-1 adapter back into the base language model and trains a fresh, smaller adapter on real-scene downstream QA. Token embeddings stay frozen across both stages.}
    \label{fig:training_stages}
\end{figure}

Given a set of $K$ posed RGB images $\{I_k\}_{k=1}^{K}$ with metric depth maps $\{D_k\}$ and pinhole camera intrinsics $C_k = (f_x, f_y, c_x, c_y)$ (per-frame focal lengths and principal point), our goal is to construct a single panoramic representation that a pretrained VLM can ingest directly. The representation is built in two steps (\Cref{fig:method}): (1)~extract per-patch features from each view and lift them into 3D world coordinates (\Cref{sec:lifting}), and (2)~reproject the lifted patches onto a shared equirectangular canvas, add a 3D position embedding of each patch's metric coordinates, and assemble the VLM's input sequence with its native 3D-RoPE position encoding (\Cref{sec:projection}). Training proceeds in two stages (\Cref{fig:training_stages}): a stage-1 spatial pretraining phase (\Cref{sec:curriculum}) followed by stage-2 adaptation on downstream target data (\Cref{sec:training}).

\subsection{Feature Extraction and 3D Lifting}
\label{sec:lifting}

Each image $I_k$ is passed through Qwen3-VL's frozen vision encoder~\cite{qwen3vl} to obtain a patch-level feature map $F_k \in \mathbb{R}^{H_f \times W_f \times D}$ at the encoder's grid resolution $H_f \times W_f$ and per-patch feature dimension $D$. Each patch $(u,v)$ is lifted to world coordinates using its depth $z = D_k(u,v)$, intrinsics scaled to the feature-map resolution, and the camera-to-world pose $T_k \in \text{SE}(3)$:
\begin{equation}
    \mathbf{p}^{\text{world}}_{u,v} = T_k \begin{pmatrix} (u - c_x)\,z/f_x \\ (v - c_y)\,z/f_y \\ z \end{pmatrix},
\end{equation}
where $T_k \in \text{SE}(3)$ acts as $T_k \mathbf{x} = R_k \mathbf{x} + \mathbf{t}_k$, and $(u, v, z)$ follow the OpenCV camera convention (x right, y down, z forward).
After lifting all $K$ views we obtain a set $\mathcal{P} = \{(\mathbf{p}_i, \mathbf{f}_i, t_i)\}$ of lifted patches, with world coordinates $\mathbf{p}_i \in \mathbb{R}^3$, features $\mathbf{f}_i \in \mathbb{R}^D$, and a per-patch temporal index $t_i \in \{1,\dots,K\}$ recording the position of the source frame in the input video. We carry $t_i$ through to panoramic placement (\Cref{sec:projection}), where it becomes the patch's coordinate on the language model's temporal MRoPE axis.

\subsection{Panoramic Canvas and Position Encoding}
\label{sec:projection}

We place each lifted patch on an equirectangular canvas defined by its origin $\mathbf{c} \in \mathbb{R}^3$ and orientation $R \in \text{SO}(3)$. The pair $(\mathbf{c}, R)$ is a free design choice and can be set to whatever frame is most natural for the task: an agent pose for situated reasoning, a viewpoint that maximizes scene coverage, or any other reference. Specific conventions used at training and inference are given in \Cref{sec:impl}.

For each lifted patch, we compute the local offset $\mathbf{q}_i = R^\top(\mathbf{p}_i - \mathbf{c}) = (q_x, q_y, q_z)$ in the canvas frame and continuous spherical coordinates: longitude $\theta_i = \operatorname{atan2}(q_x, q_z)$ and latitude $\phi_i = \arctan(-q_y/\sqrt{q_x^2 + q_z^2})$. The image y-axis points downward in the camera frame, so $q_y$ measures the downward direction and the $-q_y$ in the latitude formula gives positive latitude upward.

\paragraph{Continuous positions, not a pixel grid.} A natural alternative is to rasterize the lifted patches onto a fixed pixel grid and reduce cell contents through a hand-designed rule (averaging, depth-based selection, last-write). This collapses patches that project to the same or nearby angular position (for instance occluded surfaces along a common viewing ray or co-located observations from overlapping frames), destroying the distinction between surfaces and replacing a pretrained-attention decision with a fixed reduction. We instead keep each lifted patch as its own input token at its continuous $(\phi, \theta)$ position. Co-located patches remain distinct tokens, disambiguated by the 3D position embedding described below, and the model resolves overlap through attention rather than a pre-baked rule. Rasterization, rendered-pixel, and longitude-seam controls are reported in \Cref{app:representation_controls} and \Cref{app:seam}.

\paragraph{Position encoding.} The VLM uses its stock 3D Rotary Position Embedding (3D-RoPE)~\cite{qwen3vl} unchanged. Only the \emph{content} of the position IDs differs from the standard image/video setting. We set $W$ to the longitude $\theta_i \in [-\pi, \pi]$, $H$ to the latitude $\phi_i \in [-\pi/2, \pi/2]$, and $T$ to the source frame index $t_i \in \{1, \dots, K\}$, with all three linearly rescaled to $[0, 100]$ to stay within the effective positional distribution the model was pretrained on. $T$ retains its standard ``temporal / frame-ordering'' semantics from video. Equirectangular distortion and the longitude seam are discussed in \Cref{app:seam}.

\paragraph{3D position embedding.} A 2D VLM picks up only a class-conditioned scale prior from pixels, not a metric reading of the actual layout. We supply explicit per-patch metric position by encoding the canvas-frame offset $\mathbf{q}_i = (q_x, q_y, q_z)$ as the concatenation of (i) per-axis sinusoids $\{\sin(\omega_k q_a), \cos(\omega_k q_a), q_a\}$ for $a \in \{x, y, z\}$ at 16 log-spaced frequencies $\omega_k \in [0.1, 100]\,\text{rad/m}$, (ii) the same form on the radial primitives $\|\mathbf{q}_i\|$ and $r_{xz} = \sqrt{q_x^2 + q_z^2}$ at 8 log-spaced frequencies $\nu_k \in [0.1, 10]\,\text{rad/m}$, and (iii) the unit-ray direction $\mathbf{q}_i / \|\mathbf{q}_i\|$ as raw passthrough. These radial primitives directly expose quantities relevant to pairwise distance and horizontal-radial reads such as floor area, rather than requiring the model to synthesise them from per-axis components. The 136-channel encoding is projected by a 2-layer MLP with hidden width 64 and added to the patch feature through a learned scalar gate. Injecting metric position in feature space rather than as additional RoPE dimensions keeps 3D-RoPE's pretrained angular and temporal semantics intact, and lets the VLM attend to metric position through the same content-attention pathway it uses for any other feature.

\subsection{Spatial Pretraining Curriculum}
\label{sec:curriculum}

Spatial reasoning is difficult to acquire from downstream QA alone. The supervision is indirect, target-task distributions are narrow, and scene-appearance cues dominate the gradient before any geometric reading forms, so a model fits the per-template answer marginal long before it learns to read the input. Once that prior is absorbed it is hard to displace, since switching to a geometric readout requires a transient drop in accuracy that the QA gradient is not strong enough to reward. We therefore want supervision in which the answer follows causally from the input geometry, with no language-prior or scene-memorisation shortcut. The panoramic canvas is the natural input format for it: arbitrary 3D content can be placed at any world position with metric coordinates carried in the per-patch embedding, so the geometric answer can be read directly from the placements through the same path the model uses on a real scene.

\paragraph{Curriculum objects.} Placing recognisable real objects on the canvas can weaken the supervision we are trying to provide, since doors tend to be the same height and bathrooms a fairly consistent size, so a model that has seen enough scenes can often answer geometric questions from class identity alone. In our default formulation we populate a box-sized area of the canvas with a single feature patch sampled from a real scene, and refer to that placement in the prompt by inserting the same patch where a class name would appear. This inline copy keeps the MRoPE position of its slot in the prompt text, like any other token, rather than the longitude/latitude canvas position carried by its counterpart on the canvas. It also receives the additive 3D position embedding. Removing object identity as a shortcut makes the tasks rely primarily on the known geometry of the placements and their references. The patch is drawn from a precomputed pool of activations harvested from training-source scenes of ScanNet~\cite{dai2017scannet}, ScanNet++~\cite{yeshwanthliu2023scannetpp}, and ARKitScenes~\cite{dehghan2021arkitscenes}, so the vision encoder treats it as ordinary visual content. An alternative class-labelled scheme based on EmbodiedScan~\cite{embodiedscan} object features, and a head-to-head VSI-Bench comparison, are reported in \Cref{app:synthetic_vs_real}.

\paragraph{Task types.}  To learn geometry-based spatial understanding we construct a compact set of tasks generated on-the-fly across six families:
\begin{itemize}
    \item \textbf{Metric measurement.} Surface-to-surface distance between two placed objects, relative-distance comparisons (which of several candidates is closer to a reference, matching VSI-Bench's relative-distance task), and floor-area readout as a room-size signal.
    \item \textbf{Egocentric direction.} ``Left / right / front / back'' queries from a chosen viewpoint, at three difficulty levels, plus an $o$'clock-direction variant and a camera-perspective variant matching downstream SI-style queries.
    \item \textbf{Multi-turn navigation.} $N$-turn path queries between two placed targets, $N \in \{1,2,3,4\}$.
    \item \textbf{Observability.} When a placement first becomes visible along the frame sequence (appearance-order), and whether a target is visible from a given pose.
    \item \textbf{Counting.} Counts of placed objects under divisibility conditions (parity, divisibility by three) or location conditions (placements on the agent's left, right, front, or back).
    \item \textbf{Multi-target bounding-box readout.} Two to four placed objects, with the prompt asking for each one's 3D bounding box in canonical order.
\end{itemize}

\subsection{Two-Stage Training}
\label{sec:training}

\paragraph{Stage 1: spatial pretraining.} We train on the spatial pretraining curriculum alone, teaching the geometric reasoning skills downstream 3D-QA depends on under supervision designed to suppress the scene-prior shortcuts available in real-scene QA. Because each curriculum sample carries far fewer tokens than a full scene panorama, stage 1 is cheap to run and we can afford a long training schedule.

\paragraph{Stage 2: target adaptation.} We \emph{merge} the stage-1 LoRA weights back into the base language model and train a fresh, lower-rank LoRA adapter ($r{=}64$) on top, on a weighted mixture of downstream spatial QA (SQA3D, VSI-Bench-style data, and ViCA-family corpora). Merging freezes the stage-1 update into the base weights, and the lower rank of the new adapter limits how far stage~2 can drift from the stage-1 geometric reading while still allowing answer-format adaptation.

\section{Experiments}
\label{sec:experiments}

\subsection{Implementation Details}
\label{sec:impl}

\paragraph{Panoramic projection.} We sample $K{=}32$ frames per scene at evenly spaced timestamps along the video. The scene center is set to the annotated agent position for SQA3D and to the queried camera pose for SPBench single-image queries, and to the centroid of all camera positions otherwise. Training is run at $320{\times}240$ input resolution across all data sources. At evaluation we use $320{\times}240$ for SQA3D, where raising it gave no aggregate gain, and $640{\times}480$ for VSI-Bench and SPBench, where explicit numeric-measurement queries benefit from higher input resolution. For ScanNet scenes we crop 3\% from each image edge (with intrinsics adjusted) to avoid spurious lifted points outside the scene.  We use ground-truth depth and ground-truth camera poses whenever the dataset provides them, and fall back to estimates from a feed-forward reconstruction model on the small subset of scenes where the released ground-truth pose track is corrupted. During training only, we randomize the panorama's location and orientation per sample by drawing the scene center from the available camera positions and rotating the forward axis by a uniform random yaw.

\paragraph{Training.} We build on Qwen3-VL-8B~\cite{qwen3vl}, a vision-language model with a ViT-based vision encoder and a 3D RoPE position encoding scheme with temporal, height, and width dimensions. Training proceeds in two stages on 8 A6000 GPUs, with LoRA adapters targeting the attention and MLP projections of the language model. Stage~1 trains on the spatial pretraining curriculum with LoRA rank $r{=}256$ ($\alpha{=}512$, dropout 0.05), effective batch 16, learning rate $2{\times}10^{-5}$ with cosine schedule and 3\% warmup. The 3D position embedding trains jointly at $50{\times}$ the base learning rate. In stage~1 it is rescaled in every forward pass to half the mean norm of the visual features, so it learns direction but not magnitude. Stage~2 starts with its gate set to that same scale and then learns the scale freely. During stage~1 we cap the placed-object patch count at 200 per sample, distributed uniformly across the placed boxes with a floor of 8 patches per box to retain shape, which keeps stage~1 fast while leaving the geometry signal intact. Samples also carry 6--12 unreferenced distractor boxes, so a placement must be identified by its inline patch rather than as the only object on the canvas (\Cref{app:curriculum}). The stage-2 LoRA uses $r{=}64$, $\alpha{=}128$, dropout 0.05, and trains for 10k steps with effective batch 32 at the same learning rate, schedule, and warmup. Stage-2 training data is a weighted mixture of VLM-3R-VSIBench~\cite{vlm3r}, SQA3D~\cite{sqa3d}, and ViCA~\cite{vica}, mixed at approximately 70/15/15 by sampling weight.

\subsection{Evaluations}
\label{sec:sota}

We evaluate on three benchmarks: SQA3D~\cite{sqa3d} (situated question answering), VSI-Bench~\cite{vsibench} (eight spatial subtasks split between numerical and multiple-choice), and SPBench~\cite{spbench} (single-image and multi-view spatial reasoning, evaluated zero-shot).

\paragraph{RGB-predicted geometry.}
Without retraining, replacing ground-truth geometry with depth, poses, and intrinsics predicted from the evaluation RGB frames by Depth Anything~3~\cite{depthanything3} gives 64.9 on SQA3D, 70.0 on VSI-Bench, and 71.3 on SPBench, compared with 65.3, 71.3, and 72.1 using ground-truth geometry. The complete inference-only protocol and limitations are in \Cref{app:predicted_geometry}.

\paragraph{Spatial question answering.}

\Cref{tab:sqa3d} reports results on SQA3D, comparing our method against point-cloud-based 3D VLMs and video-input approaches. We achieve the best overall accuracy. We obtain the best results on the question types that most directly test situated reasoning (\emph{Which}, \emph{Can}, \emph{Others}), with the largest margin on \emph{Which}. The large \emph{Which} margin tracks the canvas-origin ablation (\Cref{tab:sqa3d_pose}): agent-pose centering drives most of it, and \emph{Which} is the most viewpoint-dependent question type. The remaining gap concentrates on \emph{Is} (yes/no questions, where textual priors carry most of the signal) and \emph{How} (predominantly counting, a capability the base 2D VLM is already heavily pretrained on).

\begin{table*}[!t]
    \centering
    \small
    \caption{\textbf{Evaluation on SQA3D~\cite{sqa3d} test split.}
    Comparison of methods across question types. EM@1 and EM@R1 are the primary metrics.
    Per-question-type columns report strict EM@1.
    Best results in \textbf{bold}, second best \underline{underlined}.
    Dashes indicate scores not reported by the method.
    }
    \setlength{\tabcolsep}{4pt}
    \resizebox{\textwidth}{!}{%
    \begin{tabular}{l cccccc cc}
        \toprule
        \multirow{2}{*}{Method} & \multicolumn{6}{c}{Per Question Type} & \multirow{2}{*}{EM@1} & \multirow{2}{*}{EM@R1} \\
        \cmidrule(lr){2-7}
        & What & Is & How & Can & Which & Others & & \\
        \midrule
        PQ3D~\cite{pq3d}                         & 37.1 & 61.3 & 44.5 & 60.9 & 47.0 & 45.1 & 47.1 & 49.3 \\
        3D-VisTA~\cite{3dvista}             & 34.8 & 63.3 & 45.4 & 69.8 & 47.2 & 48.1 & 48.5 & 50.9 \\
        LEO~\cite{leo}                     & 39.0 & 63.9 & 44.9 & 66.2 & 47.7 & 51.1 & 50.0 & 52.4 \\
        SIG3D~\cite{sig3d}                & 35.6 & 67.2 & 48.5 & 71.4 & 49.1 & 45.8 & 52.6 & 54.4 \\
        Scene-LLM~\cite{scenellm}        & 40.9 & 69.1 & 45.0 & 70.8 & 47.2 & 52.3 & 54.2 & 56.2 \\
        ChatScene~\cite{chatscene}               & 45.4 & 67.0 & 52.0 & 69.5 & 49.9 & 55.0 & 54.6 & 57.5 \\
        Video-3D LLM~\cite{video3dllm}           & 51.1 & 72.4 & 55.5 & 69.8 & 51.3 & 56.0 & 58.6 & 60.8 \\
        GPT4Scene-HDM~\cite{gpt4scene}           & 55.9 & 69.9 & 50.8 & 68.7 & 53.3 & 60.4 & 59.4 & 62.4 \\
        VLM-3R~\cite{vlm3r}                      & --   & --   & --   & --   & --   & --   & 60.7 & 63.4 \\
        SpaceMind~\cite{spacemind}               & 54.1 & 74.8 & \textbf{61.7} & 71.0 & 51.9 & 53.6 & 61.1 & 63.8 \\
        SR-3D~\cite{sr3d}                        & 55.0 & \underline{76.4} & 59.8 & \underline{71.6} & 54.7 & \underline{61.1} & 62.2 & -- \\
        Loc3R-VLM~\cite{qu2026loc3rvlm}          & --   & --   & --   & --   & --   & --   & 62.8 & 65.0 \\
        Ross3D~\cite{ross3d}                     & \underline{56.0} & \textbf{79.8} & \underline{60.6} & 70.4 & \underline{55.3} & 60.1 & \underline{63.0} & \underline{65.7} \\
        \midrule
        \rowcolor{gray!10}
        \textbf{\modelname{}} (Ours) & \textbf{56.6} & 74.7 & 60.2 & \textbf{75.4} & \textbf{72.6} & \textbf{65.7} & \textbf{65.3} & \textbf{68.4} \\
        \bottomrule
    \end{tabular}%
    }
    \label{tab:sqa3d}
\end{table*}

\Cref{tab:spbench} presents zero-shot results on SPBench, where every method is evaluated without SPBench training data. We top the leaderboard, with the largest gain concentrated on multi-view MCQ, consistent with the intuition that a single panoramic canvas gives multi-view questions a unified frame of reference rather than forcing the VLM to stitch several independent views. We also top the single-image split on average accuracy.

\begin{table*}[!t]
    \centering
    \small
    \caption{\textbf{Evaluation on SPBench~\cite{spbench}.}
    All models are evaluated without using SPBench training data.
    Best results per section in \textbf{bold}, second best \underline{underlined}.
    }
    \setlength{\tabcolsep}{4pt}
    \resizebox{\textwidth}{!}{%
    \begin{tabular}{l c ccc ccc}
        \toprule
        \multirow{2}{*}{Method} & \multirow{2}{*}{Overall} & \multicolumn{3}{c}{SPBench-MV} & \multicolumn{3}{c}{SPBench-SI} \\
        \cmidrule(lr){3-5} \cmidrule(lr){6-8}
        & & NQ & MCQ & Avg. & NQ & MCQ & Avg. \\
        \midrule
        Video-R1~\cite{videor1}                      & 43.8 & 32.5 & 53.0 & 42.8 & 27.7 & 62.0 & 44.9 \\
        SpaceR-7B~\cite{spacer}                    & 53.5 & 63.2 & 53.7 & 58.5 & 35.7 & 61.5 & 48.6 \\
        VILASR-7B~\cite{vilasr}                        & 54.0 & 56.2 & 59.6 & 57.9 & 36.6 & \textbf{63.7} & 50.2 \\
        Spatial-MLLM-4B~\cite{spatialmllm}             & 52.5 & 63.7 & 58.9 & 61.3 & 38.1 & 49.3 & 43.7 \\
        SpaceMind~\cite{spacemind}                      & \underline{67.3} & \underline{76.2} & \underline{70.5} & \underline{73.8} & \textbf{66.3} & 53.2 & \underline{59.7} \\
        \midrule
        \rowcolor{gray!10}
        \textbf{\modelname{}} (Ours) & \textbf{72.1} & \textbf{79.0} & \textbf{91.8} & \textbf{81.5} & \underline{62.8} & \underline{62.7} & \textbf{62.8} \\
        \bottomrule
    \end{tabular}%
    }
    \label{tab:spbench}
\end{table*}

\Cref{tab:vsibench} shows results on VSI-Bench, where we again come out on top, with the largest margin on route planning, the subtask that most directly exercises multi-step reasoning over a unified panoramic canvas. We also lead on absolute distance, relative distance, and room size, and finish second on relative direction.

\begin{table*}[!t]
    \centering
    \small
    \caption{\textbf{Evaluation on VSI-Bench~\cite{vsibench}.}
    Comparison across numerical and multiple-choice spatial reasoning subtasks.
    Best results per section in \textbf{bold}, second best \underline{underlined}.
    }
    \setlength{\tabcolsep}{4pt}
    \resizebox{\textwidth}{!}{%
    \begin{tabular}{l c cccc cccc}
        \toprule
        \multirow{2}{*}{Method} & \multirow{2}{*}{Avg.} & \multicolumn{4}{c}{Numerical Question} & \multicolumn{4}{c}{Multiple-Choice Question} \\
        \cmidrule(lr){3-6} \cmidrule(lr){7-10}
        & & Obj.Cnt. & Abs.Dist. & Obj.Size & Room Size & Rel.Dist. & Rel.Dir. & Route & App.Order \\
        \midrule
        Spacer~\cite{spacer}                       & 45.5 & 57.8 & 28.2 & 59.9 & 47.1 & 40.1 & 45.4 & 33.5 & 52.1 \\
        ViLaSR~\cite{vilasr}                           & 45.4 & 63.5 & 34.4 & 60.6 & 30.9 & 48.9 & 45.2 & 30.4 & 49.2 \\
        Spatial-MLLM~\cite{spatialmllm}                & 48.4 & 65.3 & 34.8 & 63.1 & 45.1 & 41.3 & 46.2 & 33.5 & 46.3 \\
        ViCA~\cite{vica}                               & 60.6 & 68.8 & 57.0 & \textbf{79.2} & \underline{75.1} & 58.5 & 42.6 & 34.5 & 68.8 \\
        VLM-3R~\cite{vlm3r}                           & 60.9 & 70.2 & 49.4 & 69.2 & 67.1 & 65.4 & 80.5 & 45.4 & 40.1 \\
        VST~\cite{vst}                               & 61.2 & 71.6 & 43.8 & 75.5 & 69.2 & 60.0 & 55.6 & 44.3 & 69.2 \\
        Cambrian-S~\cite{cambrians}                  & 67.5 & \underline{73.2} & 50.5 & 74.9 & 72.2 & \underline{71.1} & 76.2 & 41.8 & \textbf{80.1} \\
        SenseNova-SI~\cite{sensenovasi}              & 68.8 & 72.0 & 53.5 & 76.8 & 72.8 & 69.6 & 80.8 & \underline{48.5} & \underline{76.4} \\
        SpaceMind~\cite{spacemind}                    & \underline{69.6} & \textbf{73.3} & \underline{61.4} & \underline{77.3} & 74.2 & 67.2 & \textbf{88.4} & 44.3 & 70.6 \\
        \rowcolor{gray!10}
        \textbf{\modelname{}} (Ours) & \textbf{71.3} & 68.9 & \textbf{62.3} & 75.1 & \textbf{76.8} & \textbf{73.1} & \underline{88.0} & \textbf{60.8} & 65.7 \\
        \bottomrule
    \end{tabular}%
    }
    \label{tab:vsibench}
\end{table*}

\subsection{Ablation Studies}
\label{sec:ablations}

\begin{table*}[t]
    \centering
    \small
    \caption{\textbf{Ablation study} on VSI-Bench. Each row lists the components present in that variant. The full model uses panoramic reprojection, 3D position encoding, and stage-1 spatial pretraining, while the bottom row is the base VLM with multi-view input and none of these components. All variants are LoRA-finetuned with matched training settings. Best results per column in \textbf{bold}, second best \underline{underlined}.}
    \setlength{\tabcolsep}{4pt}
    \resizebox{\textwidth}{!}{%
    \begin{tabular}{l c cccc cccc}
        \toprule
        \multirow{2}{*}{Configuration} & \multirow{2}{*}{Avg.} & \multicolumn{4}{c}{Numerical Question} & \multicolumn{4}{c}{Multiple-Choice Question} \\
        \cmidrule(lr){3-6} \cmidrule(lr){7-10}
        & & Obj.Cnt. & Abs.Dist. & Obj.Size & Room Size & Rel.Dist. & Rel.Dir. & Route & App.Order \\
        \midrule
        \rowcolor{gray!10}
        Full model (ours)                  & \textbf{71.3} & 68.9 & \textbf{62.3} & 75.1 & \textbf{76.8} & \underline{73.1} & 88.0 & \underline{60.8} & \textbf{65.7} \\
        Full w/o 3D PE                     & \underline{70.4} & \underline{69.3} & \underline{58.1} & 76.1 & 71.3 & 70.3 & \underline{88.6} & \textbf{66.0} & \underline{63.8} \\
        Panorama + 3D PE                   & 68.5 & 68.3 & 56.6 & \textbf{76.9} & \underline{74.1} & \textbf{73.2} & \textbf{91.6} & 45.4 & 61.3 \\
        Panorama only                      & 65.3 & 68.8 & 54.8 & \underline{76.8} & 56.9 & 71.6 & 87.4 & 45.3 & 60.4 \\
        Base VLM (multi-view)              & 65.5 & \textbf{72.5} & 53.2 & 75.8 & 68.6 & 68.9 & 77.4 & 43.9 & 63.4 \\
        \bottomrule
    \end{tabular}%
    }
    \label{tab:ablation}
\end{table*}

\paragraph{Component ablation.} In \Cref{tab:ablation} we ablate the key components of our method on VSI-Bench. To match the total LoRA capacity of the two-stage full model (stage~1 $r{=}256$ merged into the base, stage~2 $r{=}64$), the single-stage ablation variants are trained with $r{=}256$ ($\alpha{=}512$). Without the spatial pretraining stage we observe a large drop on the route task, confirming the value of the curriculum for this capability. We also perform better on the tasks that require estimating metric quantities, with the exception of object size, which relies much more heavily on text priors. Removing the positional encoding produces a large additional drop on the room size task, while the other metric questions stay relatively stable as the model can still infer sizes and distances from the scene's appearance. Removing only the 3D position embedding from the full model isolates the same effect with the curriculum intact: the drop is concentrated on the metric subtasks, while angular and categorical subtasks are unaffected, consistent with the 3D position embedding acting as a metric-scale auxiliary channel on top of the panoramic canvas. We also see how much the unified panoramic representation contributes compared to passing the images straight: tasks that require relating objects across different parts of the scene perform notably better with our method, with the largest improvement on relative direction. The exception is object counting, where the base VLM retains a small lead. We attribute this to the rearrangement itself, which moves the input away from the image layout the backbone was pretrained on. Counting depends less on spatial understanding than on how many tokens show the queried object and how concentrated they are~\cite{zhang2018counting}, so the better spatial representation does not recover this cost (\Cref{app:representation_controls}).

\paragraph{Canvas origin.} We evaluate the same checkpoint on SQA3D under four canvas origin strategies (\Cref{tab:sqa3d_pose}). Using the agent's pose increases the results noticeably on most tasks, except on the \emph{Is} and \emph{How} categories, which are generally viewpoint agnostic. The random-camera origin is sampled from the raw camera positions used to construct the canvas and therefore remains inside the observed space. Only the outside-scene condition places the origin outside the room. Other camera placement strategies generally do not greatly influence the results, except for the outside-scene origin where the results drop consistently across the tasks.

\begin{table}[t]
    \centering
    \small
    \caption{\textbf{Canvas origin ablation on SQA3D.}
    Same checkpoint evaluated under four canvas origin strategies. \emph{Agent pose} tracks the agent's position and heading. The other three fix forward to world $+X$ with origins at the scene centroid, a raw camera position inside the observed space, or outside the scene.
    Per-question-type columns report strict EM@1.
    Best per column in \textbf{bold}, second best \underline{underlined}.
    }
    \setlength{\tabcolsep}{5pt}
    \resizebox{0.8\textwidth}{!}{%
    \begin{tabular}{l cccccc cc}
        \toprule
        \multirow{2}{*}{Canvas origin} & \multicolumn{6}{c}{Per Question Type} & \multirow{2}{*}{EM@1} & \multirow{2}{*}{EM@R1} \\
        \cmidrule(lr){2-7}
        & What & Is & How & Can & Which & Others & & \\
        \midrule
        \rowcolor{gray!10}
        Agent pose      & \textbf{56.6} & \textbf{74.7} & \underline{60.2} & \textbf{75.4} & \textbf{72.6} & \textbf{65.7} & \textbf{65.3} & \textbf{68.4} \\
        Scene center    & 54.6 & 74.2 & 59.8 & \underline{68.9} & \underline{56.7} & \underline{58.9} & 61.2 & 64.3 \\
        Random camera   & \underline{54.8} & \underline{74.4} & \textbf{61.3} & \underline{68.9} & 55.6 & \underline{58.9} & \underline{61.4} & \underline{64.6} \\
        Outside scene   & 54.0 & 73.6 & 60.0 & 68.3 & 51.3 & 55.7 & 59.8 & 62.7 \\
        \bottomrule
    \end{tabular}%
    }
    \label{tab:sqa3d_pose}
\end{table}

\section{Conclusion}
\label{sec:conclusion}

\modelname{} reframes 3D scene understanding as 2D spatial reasoning on a panoramic canvas. Multi-view RGB-D is reprojected onto a single equirectangular panorama with each patch carrying its metric 3D position through an additive 3D position embedding, and the pretrained VLM reads the result through its native attention. A spatial pretraining curriculum of objects placed on an otherwise empty canvas and referenced from the prompt warms up geometric reading before downstream fine-tuning. The combination sets a new state of the art on VSI-Bench and SQA3D, and tops the zero-shot SPBench leaderboard. The representation also carries over to other backbones (\Cref{app:backbones}).

\paragraph{Limitations.}
\modelname{} requires depth and camera poses, which pure RGB methods avoid, and pose-estimation failures can still degrade the canvas, though the inference-only Depth Anything~3 evaluation in \Cref{app:predicted_geometry} narrows this gap. The single global panorama trades fine spatial precision for compactness and may limit very large or outdoor scenes. The pretraining curriculum is hand-authored, so adding a qualitatively new spatial skill requires writing a new task generator rather than collecting more data.

\begin{ack}
This project was funded by the ERC Consolidator Grant Gen3D (101171131).
We also thank Angela Dai for the video voice-over.
\end{ack}

\bibliographystyle{plainnat}
\bibliography{references}

\appendix

\section{Object Referencing: Alternative Class-Labelled Design}
\label{app:synthetic_vs_real}

The main paper places \emph{synthetic axis-aligned boxes} on the canvas whose per-patch features are drawn from a shared random pool of activations harvested from training-source scenes, and references each placement in the prompt via an inline copy of the same patch feature. The motivation is to disconnect object size and appearance from class identity, so geometry on the canvas becomes the principal signal for solving the curriculum (see the curriculum-objects discussion in the main method).
A natural alternative is to populate the canvas with real objects harvested in-context from scene instance annotations~\cite{embodiedscan}, pasted onto the canvas with their original per-patch features, and referenced in the prompt by class name. Features are extracted within the original scene rather than re-encoded from cropped objects in isolation, since the latter would discard the surrounding-scene context each patch was encoded with. This variant adds text-to-object grounding as a side benefit, at the cost of re-introducing the class priors (typical door height, typical bathroom size) that the synthetic-box scheme is designed to suppress.
\Cref{tab:vsi_synthetic_vs_real} compares the two curriculum content choices on the full VSI-Bench test split, holding the backbone, canvas construction, MRoPE schedule, task families, family-flat weighting, stage-2 fine-tuning, and evaluation protocol fixed. The only thing that changes between rows is what populates the canvas in stage~1.

\begin{table}[h]
    \centering
    \small
    \caption{\textbf{Curriculum content study on VSI-Bench (full test split).}
    Both rows share the same backbone, canvas construction, task families, stage-2 fine-tuning, and evaluation protocol. Only the stage-1 curriculum content differs. \emph{Synthetic boxes}: axis-aligned boxes with shared-random per-patch features, referenced in the prompt by inline patch. \emph{Scene-harvested objects}: real objects pasted from in-context scene crops with their original per-patch features, referenced by class name.}
    \setlength{\tabcolsep}{4pt}
    \resizebox{\textwidth}{!}{%
    \begin{tabular}{l c cccc cccc}
        \toprule
        \multirow{2}{*}{Curriculum content} & \multirow{2}{*}{Avg.} & \multicolumn{4}{c}{Numerical Question} & \multicolumn{4}{c}{Multiple-Choice Question} \\
        \cmidrule(lr){3-6} \cmidrule(lr){7-10}
        & & Obj.Cnt. & Abs.Dist. & Obj.Size & Room Size & Rel.Dist. & Rel.Dir. & Route & App.Order \\
        \midrule
        Synthetic boxes (main paper)   & \textbf{71.3} & 68.9 & \textbf{62.3} & 75.1 & 76.8 & \textbf{73.1} & \textbf{88.0} & \textbf{60.8} & \textbf{65.7} \\
        Scene-harvested objects        & 70.2 & \textbf{69.8} & 61.1 & \textbf{76.5} & \textbf{79.6} & 67.8 & 86.1 & 59.8 & 60.8 \\
        \bottomrule
    \end{tabular}%
    }
    \label{tab:vsi_synthetic_vs_real}
\end{table}

\section{Additional Experiments and Analyses}
\label{app:rebuttal_experiments}

This section collects additional evaluations. The protocols are not interchangeable. The predicted-geometry, rasterization, and seam studies change only inference. The feature-versus-pixel and ERNIE comparisons are matched 2{,}000-step training runs. The seed replications use the complete 10{,}000-step schedule. The Qwen3.5 result uses the full two-stage recipe. We state these boundaries with each result.

\subsection{Geometry Predicted from RGB}
\label{app:predicted_geometry}

We replace all ground-truth depth, poses, and intrinsics with estimates produced directly from RGB by Depth Anything~3 (DA3NESTED-GIANT-LARGE-1.1)~\cite{depthanything3}. Reconstruction uses exactly the frames supplied to the VLM at evaluation: $K{=}32$ for SQA3D and VSI-Bench, and the number of views supplied with each SPBench question. It uses no extra views and no ground-truth geometry. This is an inference-only substitution using the paper checkpoint, with no retraining around the geometry estimator.

\begin{table}[h]
    \centering
    \small
    \caption{\textbf{Feed-forward predicted geometry.} Accuracy under the paper's full evaluation protocol. The previous-best row names the strongest prior result separately on each benchmark: Ross3D on SQA3D and SpaceMind on VSI-Bench and SPBench.}
    \setlength{\tabcolsep}{7pt}
    \begin{tabular}{l r r r}
        \toprule
        Geometry & SQA3D & VSI-Bench & SPBench \\
        \midrule
        Previous best & 63.0 & 69.6 & 67.3 \\
        Ours, ground-truth geometry & \textbf{65.3} & \textbf{71.3} & \textbf{72.1} \\
        Ours, RGB-predicted geometry & 64.9 & 70.0 & 71.3 \\
        \bottomrule
    \end{tabular}
    \label{tab:predicted_geometry}
\end{table}

Predicted geometry reduces accuracy by 0.4, 1.3, and 0.8 points, respectively, while exceeding the prior-method results listed in the table on all three benchmarks. This establishes robustness to this specific feed-forward estimator, not invariance to reconstruction failure or the performance of a jointly trained geometry encoder. Because the estimator is called as a replaceable preprocessing module rather than trained jointly with the VLM, the result also does not require adapting the paper checkpoint to DA3.

\subsection{Continuous Features, Rasterization, and Rendered Pixels}
\label{app:representation_controls}

\paragraph{Inference-only rasterization.}
We bin lifted patches into a fixed equirectangular grid, mean-pool the features, depth, and frame index of all patches in a cell, and assign the pooled token the cell-center angle. The $181{\times}362$, $90{\times}180$, and $45{\times}90$ grids have approximately square angular cells of $1^\circ$, $2^\circ$, and $4^\circ$. All rows use the paper evaluation protocol and the same checkpoint trained on the continuous canvas. The rasterized scores are therefore inference-only ablations. A model retrained for a grid could perform differently and was not evaluated.

\begin{table}[h]
    \centering
    \small
    \caption{\textbf{Inference-only rasterization on VSI-Bench.} Token counts are the occupied-cell counts presented to the model.}
    \setlength{\tabcolsep}{9pt}
    \begin{tabular}{l r r}
        \toprule
        Canvas & Tokens & VSI-Bench \\
        \midrule
        Continuous (ours) & 9{,}600 & \textbf{71.1} \\
        Rasterized $181{\times}362$ & 7{,}144 & 68.2 \\
        Rasterized $90{\times}180$ & 4{,}009 & 67.7 \\
        Rasterized $45{\times}90$ & 1{,}563 & 65.8 \\
        \bottomrule
    \end{tabular}
    \label{tab:rasterization}
\end{table}

Most of the loss appears at the finest grid. Appearance Order falls from 62.6 to 39.8 at $181{\times}362$, accounting for 2.8 points of the 2.9-point average gap. Replacing mean pooling with one original patch per occupied cell keeps the grid and token count fixed and scores 68.6 rather than 68.2. Thus averaging itself does not explain the gap. Both reductions retain only one token and one frame index per cell: repeated observations of a surface either have their frame indices averaged or all but one discarded, so first appearance is no longer recoverable. A raster grid may nevertheless extend the canvas to VLMs whose 1D positional encoding cannot assign arbitrary angular coordinates.

\paragraph{Matched feature-versus-pixel training.}
We also train two arms for 2{,}000 steps using the same base checkpoint, LoRA rank 256, effective batch 8, schedule, data mixture, and held-out VSI-Bench questions. They differ only in scene input: projected per-frame features versus a pre-rendered equirectangular z-buffer panorama.

\begin{table}[h]
    \centering
    \small
    \caption{\textbf{Feature canvas versus rendered pixels.} Matched 2{,}000-step training runs on VSI-Bench. These are short-run comparisons, not the paper's full training schedule.}
    \setlength{\tabcolsep}{10pt}
    \begin{tabular}{l r}
        \toprule
        Scene input & VSI-Bench \\
        \midrule
        Feature canvas (ours) & \textbf{60.8} \\
        Pixel panorama & 52.1 \\
        \bottomrule
    \end{tabular}
    \label{tab:features_pixels}
\end{table}

The 8.7-point difference is larger than the roughly one-point step-to-step variation observed in either run. It jointly reflects two differences and should not be read as isolating either one: z-buffer rendering collapses occluded and co-located surfaces to one surface per direction, and the rendered arm asks a vision tower pretrained on perspective images to re-encode an equirectangular image. Our canvas applies no visibility test, keeps these observations as separate tokens, and extracts their appearance features from the original perspective frames. It also permits changing the canvas origin without rendering again.

\paragraph{Counting and depth coarsening.}
The base VLM retains a small lead on object counting in \Cref{tab:ablation}. A depth-coarsening control does not support the explanation that reprojection inaccuracies fragment or merge instances. Coarsening by up to $8\times$ produces a median patch displacement of 0.16 m and moves 29\% of patches to a different surface, while counting scores remain similar across the four levels (68.4, 68.8, 69.3, and 68.9). We instead attribute the gap to the rearrangement itself, which moves the input away from the backbone's pretrained image layout, while counting depends mainly on how many tokens show the queried object and how concentrated they are~\cite{zhang2018counting}.

\subsection{Training-Seed Replications}
\label{app:seeds}

We trained three additional runs under different random seeds with every other setting fixed, retained all runs, and used the complete 10{,}000-step schedule rather than a rebuttal-window short run. As throughout the paper, checkpoints are selected by held-out generation accuracy.

\begin{table}[h]
    \centering
    \small
    \caption{\textbf{Seed variance under full 10{,}000-step training.} Standard deviations are across the three reported seeds.}
    \setlength{\tabcolsep}{8pt}
    \begin{tabular}{l r r r}
        \toprule
        & VSI-Bench & SQA3D (EM@1) & SPBench \\
        \midrule
        Seed A & 71.0 & 66.2 & 73.4 \\
        Seed B & 69.9 & 64.5 & 72.8 \\
        Seed C & 70.3 & 64.7 & 73.4 \\
        Mean $\pm$ std. & $70.4 \pm 0.6$ & $65.1 \pm 0.9$ & $73.2 \pm 0.3$ \\
        \bottomrule
    \end{tabular}
    \label{tab:seed_variance}
\end{table}

All three runs exceed the prior best of 69.6 on VSI-Bench, 63.0 on SQA3D, and 67.3 on SPBench. These three seeds quantify training-seed variation under one fixed recipe. They do not capture variation from data, hyperparameter, or model-family choices.

\subsection{Additional Backbones}
\label{app:backbones}

\paragraph{ERNIE-4.5-VL-28B-A3B.}
We ported the representation outside the Qwen family. ERNIE runs end to end on real weights through canvas prefill, training, and generation, with the adapter confined to the language model, the vision tower unchanged, and no architectural change. The two arms below differ only in whether they use the canvas and are trained for 2{,}000 iterations with identical data, schedule, adapter shape, and effective batch.

\begin{table*}[h]
    \centering
    \scriptsize
    \caption{\textbf{Matched 2{,}000-step ERNIE-4.5-VL-28B-A3B comparison on VSI-Bench.} This is a stage-2-only short run without the spatial curriculum, so the controlled row difference, rather than the absolute accuracy or a comparison to the fully trained paper model, is the relevant quantity.}
    \setlength{\tabcolsep}{3.2pt}
    \resizebox{\textwidth}{!}{%
    \begin{tabular}{l r r r r r r r r r}
        \toprule
        & Obj. Cnt. & Abs. Dist. & Obj. Size & Room Size & Rel. Dist. & Rel. Dir. & Route & App. Order & Overall \\
        \midrule
        Backbone alone & 50.6 & 26.9 & 70.5 & 42.2 & \textbf{52.3} & 44.0 & 21.3 & 13.9 & 40.2 \\
        + canvas & \textbf{65.2} & \textbf{27.9} & \textbf{71.2} & \textbf{55.8} & 51.3 & \textbf{46.8} & \textbf{28.0} & \textbf{24.6} & \textbf{46.3} \\
        $\Delta$ & +14.6 & +1.0 & +0.7 & +13.6 & -1.0 & +2.8 & +6.7 & +10.7 & +6.1 \\
        \bottomrule
    \end{tabular}%
    }
    \label{tab:ernie_backbone}
\end{table*}

ERNIE is a general-purpose model rather than a spatial specialist, and these runs omit stage~1. The +6.1 overall difference therefore supports portability of the representation under this matched short protocol, but does not establish ERNIE's full-recipe performance.

\paragraph{Qwen3.5-9B.}
Using the full two-stage recipe and the paper's full evaluation protocol, Qwen3.5-9B reaches 69.42 on VSI-Bench, 67.18 EM@1 on SQA3D, and 69.81 on SPBench. This is a full-scale run with a newer Qwen generation and different vision stack, not the requested non-Qwen control. It is reported separately from the ERNIE comparison and does not replace the paper checkpoint or its 71.3, 65.3, and 72.1 main results.

\subsection{Distortion and the Longitude Seam}
\label{app:seam}

Patch features come from the original perspective frames, so the equirectangular mapping changes only where a patch is placed, not its features. Longitude is periodic, but its assigned position ID is not: patches at $+179^\circ$ and $-179^\circ$ are nearby in the scene but receive opposite ends of the longitude index range. The additive 3D position embedding remains continuous because it encodes each patch's Cartesian offset from the canvas origin, and a uniformly random training yaw moves the seam across longitudes rather than systematically splitting one scene region. A localized angular-index discontinuity nevertheless remains.

We isolate it with an inference-only pairwise-distance probe. The two objects and answer remain fixed while the canvas orientation changes so that the seam lies either between the objects or away from them. Depth co-varies with angular separation $\Delta$ by construction, so each row is compared only internally.

\begin{table}[h]
    \centering
    \small
    \caption{\textbf{Longitude-seam probe.} Mean absolute distance error in metres.}
    \setlength{\tabcolsep}{8pt}
    \begin{tabular}{r r r}
        \toprule
        $\Delta$ & Seam not between & Seam between \\
        \midrule
        $15^\circ$ & 0.14 & 1.23 \\
        $30^\circ$ & 0.15 & 0.50 \\
        $60^\circ$ & 0.24 & 0.30 \\
        $170^\circ$ & 0.69 & 0.71 \\
        \bottomrule
    \end{tabular}
    \label{tab:seam}
\end{table}

The increase concentrates where the seam separates objects with small angular separation and becomes negligible as separation grows. Placing both objects immediately beside the seam without separating them gives 0.23~m error at $15^\circ$, isolating the effect to the wrap rather than proximity to the boundary. Under random canvas orientation this configuration affects 0.63\% of VSI-Bench absolute-distance questions (one of 214 in our evaluation). At inference, the free canvas yaw can rotate the seam away from a region of interest. Settings where seam-straddling queries are common may instead require targeted curriculum examples or a wrap-aware positional encoding.

\section{Training Compute Budget}
\label{app:compute}

The reproducible two-stage \modelname{} recipe uses 8 NVIDIA RTX A6000 GPUs (Ampere architecture, 48~GB GDDR6 each) with DeepSpeed ZeRO-2 and Flash Attention~2. Stage~1 (spatial pretraining) takes $\approx$25.2~h of wall-clock time, and Stage~2 (QA fine-tuning) takes $\approx$39.6~h, for a total of $\approx$64.8~h on 8 GPUs (518 raw A6000-GPU-hours). \Cref{tab:compute} compares this with the training costs reported by competing methods through an analytic A100-equivalent normalization. These are not measured cross-hardware runtime comparisons.

\begin{table}[h]
    \centering
    \small
    \caption{\textbf{Training compute comparison.}
    Wall-clock training time and total compute for \modelname{} and competing methods, analytically normalized to A100-equivalent GPU-hours using the per-GPU-type peak-throughput factors in \Cref{tab:gpu_norm}.
    \modelname{} numbers combine both training stages.}
    \setlength{\tabcolsep}{5pt}
    \begin{tabular}{l c r r r r}
        \toprule
        Method & GPU type & \# GPUs & Wall-clock & Raw GPU-h & A100-equiv.\ GPU-h \\
        \midrule
        \textbf{\modelname{} (Ours)}    & A6000              & 8   & $\approx$64.8~h & 518       & 259 \\
        VLM-3R~\cite{vlm3r}             & H200               & 16  & 5~h           & 80          & 240 \\
        ViCA~\cite{vica}                & H100               & 8   & 55~h          & 440         & 1{,}320 \\
        SpaceMind~\cite{spacemind}      & H100               & 64  & 25~h          & 1{,}600     & 4{,}800 \\
        SenseNova-SI~\cite{sensenovasi} & H100$^{\dagger}$   & 128 & 72~h          & 9{,}216     & 27{,}648 \\
        \bottomrule
    \end{tabular} \\[2pt]
    {\footnotesize $^{\dagger}$ GPU type not stated in the paper, assumed H100.}
    \label{tab:compute}
\end{table}

\paragraph{GPU normalization.}
We convert raw GPU-hours to A100-equivalent GPU-hours by multiplying by the dense BF16 Tensor Core throughput ratio between the source GPU and the A100~SXM4. We use \emph{dense} BF16 (no structured 2:4 sparsity) because that matches what training actually runs: BF16 numerics, no sparsity kernels. The factors are listed in \Cref{tab:gpu_norm} and are simply $\text{TFLOPS}_{\text{src}} / \text{TFLOPS}_{\text{A100}}$ rounded to one significant digit. This analytic conversion captures peak compute only and should be read as a normalized upper bound, not a measured speed ratio. In practice the achievable speedup on H100/H200 is bounded below the $3\times$ ratio by memory bandwidth (HBM ratio $1.67\times$ for H100, $2.4\times$ for H200) and by multi-node scaling efficiency at the 64- and 128-GPU jobs in \Cref{tab:compute}. Under the more conservative bandwidth-bound conversion, SpaceMind and SenseNova-SI still use approximately $10\times$ and $59\times$ our normalized compute, respectively.

\begin{table}[h]
    \centering
    \small
    \caption{\textbf{GPU normalization factors used in \Cref{tab:compute}.} Dense BF16 Tensor Core TFLOPS (no 2:4 sparsity) are NVIDIA datasheet values, and HBM bandwidth is shown for reference but is not used in the conversion. The A100-equivalent factor is computed as $\text{TFLOPS}_{\text{src}} / \text{TFLOPS}_{\text{A100}} = \text{TFLOPS}_{\text{src}} / 312$, rounded to one significant digit.}
    \setlength{\tabcolsep}{6pt}
    \begin{tabular}{l r r c}
        \toprule
        GPU & Dense BF16 & HBM bandwidth & A100-equiv.\ factor \\
        \midrule
        A6000              & 154.8~TFLOPS & 768~GB/s     & $154.8 / 312 \approx 0.5\times$ \\
        A100~SXM4          & 312~TFLOPS   & 2{,}000~GB/s & $1.0\times$ \\
        H100~SXM5          & 989~TFLOPS   & 3{,}350~GB/s & $989 / 312 \approx 3.0\times$ \\
        H200~SXM5          & 989~TFLOPS   & 4{,}800~GB/s & $989 / 312 \approx 3.0\times$ \\
        \bottomrule
    \end{tabular}
    \label{tab:gpu_norm}
\end{table}

The H200 shares the GH100 compute die with the H100 and differs only in memory capacity (141~GB HBM3e vs.\ 80~GB HBM3) and bandwidth, so it carries the same dense BF16 throughput and the same conversion factor. The 989/312 ratio rounds down from $3.17$ to $3.0$, which mildly understates competitor compute and therefore biases the comparison \emph{against} our method.

\paragraph{Per-method calculations.}
\Cref{tab:compute_calc} applies the factors from \Cref{tab:gpu_norm} to the wall-clock training time reported by each method.

\begin{table}[h]
    \centering
    \small
    \caption{\textbf{Per-method A100-equivalent compute.} A100-equivalent GPU-hours equal $(\text{\# GPUs}) \times (\text{wall-clock hours}) \times (\text{A100-equiv.\ factor})$. \# GPUs and wall-clock are taken from each method's paper, factors come from \Cref{tab:gpu_norm}.}
    \setlength{\tabcolsep}{6pt}
    \begin{tabular}{l r r r r r}
        \toprule
        Method & \# GPUs & Wall-clock (h) & Factor & A100-equiv.\ GPU-h & vs.\ Ours \\
        \midrule
        \textbf{\modelname{} (Ours)}    & 8   & 64.8 & $0.5\times$ & 259 & $1\times$ \\
        VLM-3R~\cite{vlm3r}             & 16  & 5    & $3\times$   & 240          & $0.9\times$ \\
        ViCA~\cite{vica}                & 8   & 55   & $3\times$   & 1{,}320      & $5.1\times$ \\
        SpaceMind~\cite{spacemind}      & 64  & 25   & $3\times$   & 4{,}800      & $19\times$ \\
        SenseNova-SI~\cite{sensenovasi} & 128 & 72   & $3\times$   & 27{,}648     & $107\times$ \\
        \bottomrule
    \end{tabular}
    \label{tab:compute_calc}
\end{table}

The compute claim is scoped to methods near our accuracy, not to every baseline. SpaceMind (69.6 VSI-Bench) and SenseNova-SI (68.8) are the two closest methods to our 71.3 and use $19\times$ and $107\times$ the peak-normalized compute above. VLM-3R uses slightly less normalized compute than ours (240 versus 259 A100-equivalent GPU-hours), and ViCA uses $5.1\times$ ours, but their VSI-Bench scores are 60.9 and 60.6. At comparable compute, VLM-3R is 10.4 points lower. Thus the evidence supports lower training cost at comparable accuracy, not a universal compute advantage.

\section{Spatial Pretraining Curriculum: Task Details}
\label{app:curriculum}

This section details the tasks that make up the stage-1 spatial pretraining curriculum. The curriculum organizes its task instances into six families and applies family-flat weighting: each family contributes an equal share of every minibatch, and every task within a family carries the same weight. The six families are Metric Measurement, Egocentric Direction, Multi-Turn Navigation, Observability, Counting, and Multi-Target Bounding-Box Readout.

All tasks are trained on a canvas containing only the placed objects, at procedurally sampled centres, dimensions, and yaws. No scene is ever loaded onto the canvas during stage 1, so the model only sees the placed objects.

Each placed box is realised as the set of canvas patches that fall within its volume, all carrying a single feature vector sampled from a precomputed pool of real ViT patch features. The pool is built once by running Qwen3-VL's frozen vision encoder on training-source scenes from ScanNet~\cite{dai2017scannet}, ARKitScenes~\cite{dehghan2021arkitscenes}, and ScanNet++~\cite{yeshwanthliu2023scannetpp}, and box features are drawn independently from it at training time. Because the pool is scene-agnostic, a box cannot leak the current scene's identity, and multi-target tasks cannot shortcut by clustering on tokens from a shared source frame. Because the patches are real ViT outputs, their per-token feature distribution matches what the VLM sees on real scenes, so the stage-1 reading capability transfers to stage 2 where the canvas is populated with real-scene patches.

\paragraph{Distractor boxes.} Samples add 6--12 boxes that the prompt never references. Each draws its own feature, disjoint from the task referents and from the other distractors, so neither the odd one out nor a coherent class is a usable shortcut. Floor-area tasks are sampled without distractors, since extra boxes would change the quantity being asked about.

We describe each family below, giving each task a short descriptive name followed by its precise input and output format.

\subsection{Metric Measurement}

\paragraph{Pairwise distance.} Two boxes are placed on the canvas with non-overlapping supports and a surface-to-surface separation of at least 5~cm. The prompt references each via an inline patch, and the model is asked to report the shortest distance between the two surfaces in metres, rounded to one decimal. The label is the scalar regression target.

\paragraph{Closest-of-four.} Five boxes are placed on the canvas (one \emph{target}, four \emph{candidates}), with the winner-vs.-runner-up surface-distance gap constrained to at least 5~cm. The prompt references the target and the four candidates via inline patches, and the model picks which of the four candidates is closest to the target as a 4-way MCQ with answers \{A, B, C, D\}.

\paragraph{Non-rectangular floor area.} A synthetic L- or T-shaped floor polygon is placed in the scene, rotated by a random yaw, and rendered as a thin slab of lifted points. The model reads off the floor area in square metres (one-decimal regression).

\subsection{Egocentric Direction}

The world-frame direction tasks (the two-way, three-way, four-way diagonal, four-way cardinal, and clock-face variants below) place 3 boxes: a ``ref'' box at the agent position, a ``fwd'' box defining the facing direction, and a target box to classify. The prompt reads ``If you are standing at [ref] and facing [fwd], $\ldots$'', with the bracketed names rendered as inline patches bound to the corresponding canvas patches. The camera-frame direction tasks place 2 boxes (target and pivot), with the canvas itself reoriented onto a real scene camera pose so no explicit facing reference is needed. The answer space varies by task.

\paragraph{Two-way direction.} 2-way MCQ: \{left, right\}. The question is ``is the target to the left or the right?''.

\paragraph{Three-way direction.} 3-way MCQ: \{left, right, back\}. The ``back'' bin activates when the target lies more than $135^\circ$ from the heading.

\paragraph{Four-way diagonal direction.} 4-way diagonal MCQ: \{front-left, front-right, back-left, back-right\}. Sampling enforces an angular margin from the cardinal axes so the diagonal label is well-defined.

\paragraph{Four-way cardinal direction.} 4-way cardinal MCQ \{front, back, left, right\} with $\pm 45^\circ$ bin boundaries.

\paragraph{Camera-frame direction (easy / medium / hard).} Same three answer spaces as the world-frame two-way, three-way, and four-way diagonal variants, but the canvas is recentered and reoriented onto a real scene camera pose drawn from the frame sequence. The question is phrased from the camera's perspective, ``From the camera's perspective, is X to Y's \ldots ?''.

\paragraph{Clock-face bearing.} The answer is a clock-face bearing in $\{1, 2, \dots, 12\}$, computed as the clockwise angle from the heading (12 = forward, 3 = right, 6 = back, 9 = left). Finer angular resolution than the cardinal or diagonal variants.

\subsection{Multi-Turn Navigation}

The Navigation family carries the N-turn route-planning task at four turn counts ($N \in \{1, 2, 3, 4\}$), each carrying equal weight in the family.

\paragraph{N-turn route planning ($N \in \{1, 2, 3, 4\}$).} A start box, a goal box, and $N$ intermediate waypoint boxes are placed at sampled positions on the canvas. The prompt references each via inline patches. The answer is the sequence of egocentric turn actions that takes the agent from start to goal, with one action per turn. The format is a fill-in MCQ where the question lists the actions as ``[please fill in]'' slots and the model selects one of 4 letter options. Each path-turn is drawn from \{Turn Left, Turn Right\}. Additionally, on the first turn we add a ``Turn Back'' option, so the initial move is drawn from \{Turn Left, Turn Right, Turn Back\}.

\subsection{Observability}

\paragraph{Appearance order.} Four boxes are placed at sampled positions on the canvas. Each box's per-patch MRoPE $T$ values are overridden to a controlled distribution: the four boxes are assigned consecutive first-appearance $T$-starts $\{k, k+1, k+2, k+3\}$ for a random $k$, and each box's remaining patch $T$ values are sampled uniformly in $[t_{\text{start}}, T_{\max}]$ with one patch pinned to $t_{\text{start}}$. The prompt references the four via inline patches and asks for the full first-time appearance order as a 4-way MCQ (correct permutation plus three random-permutation distractors). The construction isolates fine-grained $T$-axis resolution at the one-frame scale, with no scene content available to shortcut the answer.

\paragraph{Line-of-sight visibility.} A yes/no task: given a viewer box $p_1$, a target box $p_2$, and a synthetic axis-aligned occluder box between them, does the line of sight from $p_1$ reach $p_2$? Visibility is computed from first principles as the fraction of $p_2$'s AABB reachable from $p_1$ by a segment not intersecting the occluder, and the label is ``yes'' if that fraction is at least 10\%. The occluder is always present so box count cannot leak the label. Blocker side, along-segment position, and segment length are drawn from a shared distribution across both labels, with only the perpendicular offset of the occluder centre label-conditional. The task is \emph{not} temporal. The features are pose-independent, and the task is listed under Observability because it tests the same ``can the model resolve whether a region is reachable'' capability as appearance order.

\subsection{Counting and Arithmetic Readout}

\paragraph{Object counting.} $N$ boxes are placed on the canvas with $N$ drawn uniformly over a wide integer range. All $N$ boxes share their feature draw with a single reference box that the prompt references via an inline patch: ``How many [ref] are in the room?''. The model emits the integer count. The shared feature gives the $N$ boxes a common identity that the model can read through content attention, while non-referenced distractor boxes carry their own different feature draws so they cannot be mistaken for the referent.

\paragraph{Count parity.} Same setup as object counting but the question is ``Is the number of [ref] in the room odd or even?''. The answer is ``even'' or ``odd''. The placed count is sampled to make the two labels equiprobable, so the model cannot solve the task by guessing the marginal.

\paragraph{Count divisibility-by-three.} Same setup but the question is ``Is the number of [ref] in the room a multiple of three?''. The answer is ``yes'' or ``no''. The placed count is sampled to make the two labels equiprobable. Phrased as a binary divisibility check rather than a three-way residue classification, which keeps the answer format consistent with the other yes/no tasks.

\paragraph{Count-by-side.} A hybrid task. Given an origin pose with a forward heading and a reference box, the model counts how many same-feature boxes lie on a specified cardinal side from the agent's perspective. The side is drawn uniformly from \{right, left, front, back\} and the target count is drawn from $\{3, 4, 5, 6\}$ with weights $[30, 35, 25, 10]$. The answer is the integer. The task combines direction classification with counting.

\subsection{Multi-Target Bounding-Box Readout}

\paragraph{Multi-target bounding-box readout.} Two to four boxes are placed on the canvas at sampled centres, dimensions, and yaws. The prompt references each via an inline patch tagged with a letter: ``Provide the 3D bounding box of each highlighted region, in the same order: A: [m1], B: [m2], $\ldots$''. The model emits the axis-aligned 3D bounding box (centre and metric extents) for each box in canonical near-to-far order.

\end{document}